\documentclass[runningheads]{llncs}

\usepackage{subcaption}
\usepackage{amsmath,amssymb,amsfonts}
\usepackage{graphicx}
\usepackage{multirow}
\usepackage{multicol}
\usepackage{textcomp}
\usepackage{gensymb}
\usepackage[normalem]{ulem}
\usepackage{placeins}

\begin{document}

\title{Ankle Joints Are Beneficial When Optimizing Supported Real-world Bipedal Robot Gaits}
\titlerunning{Ankle Joints Are Beneficial for Supported Bipedal Robots}








\author{Hilmar Elverhøy\inst{1} \and
Steinar Bøe\inst{1} \and
Vegard Søyseth\inst{1} \and
Tønnes Nygaard\inst{1, 2}}

\authorrunning{H. Elverhøy et al.}

\institute{University of Oslo, Oslo, Norway \and
Norwegian Defence Research Institute, Oslo, Norway
\email{\{hilmare,steinboe,vegardds,tonnesfn\}@ifi.uio.com}}

\maketitle

\begin{abstract}

Legged robots promise higher versatility and the ability to traverse much more difficult terrains than their wheeled counterparts.
Even though the use of legged robots have increased drastically in the last few years, they are still not close to the performance seen from legged animals in nature.
Robotic legs are typically fairly simple mechanically, and few feature an ankle joint, even though most land mammals have one.
The ankle could be a key to better performance and stability for legged robots, and in this paper we investigate how the use of an ankle in a bipedal robot could improve its performance when combined with evolutionary techniques for gait optimization.
Our study shows, both in simulation and physical experiments, that the addition of an ankle joint results in greater walking speeds for a supported bipedal robot.





\end{abstract}

\section{Introduction}

\begin{figure}[t]
  \centering
  \includegraphics[height=4.3cm]{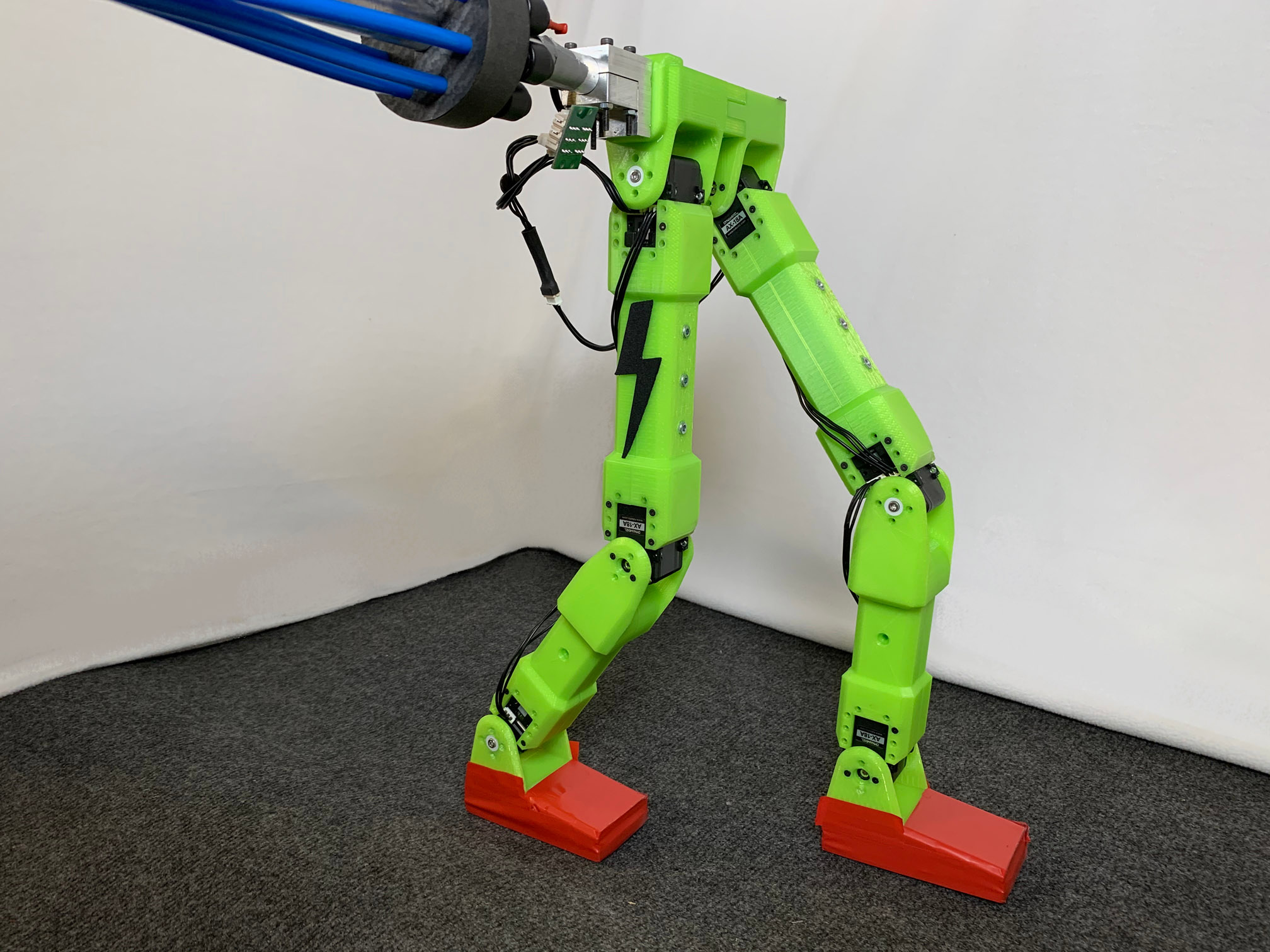}
  \hspace{2mm}
  \includegraphics[height=4.3cm]{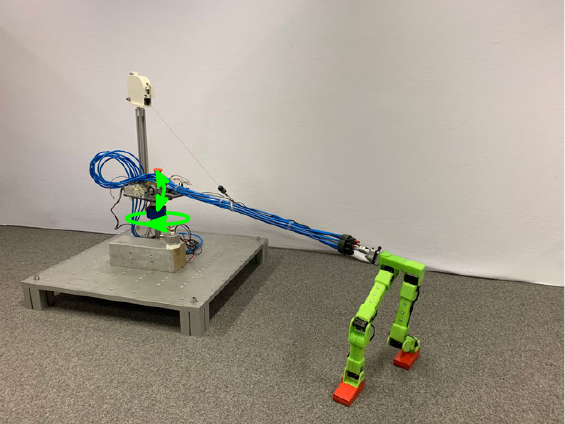}
  \caption{The physical robot used in our experiments shown to the left, with the hub setup used in the physical experiments shown to the right.}
  \label{fig.robot}
  \vspace{-5mm}
\end{figure}

In both human and animal anatomy, the ankle joint has an important role in stability and speed. Few bipedal robots have ankle joints and even fever use ankle joints actively in their gait. The function of the ankle joint is studied thoroughly in biomechanics and classical mechatronics. However, the effect of active ankle joints in bipedal robots combined with gait optimization through machine learning techniques is fairly unexplored.


Inspired by the mammal anatomy, we have explored how ankle joints affect performance in a legged robot.
We built a supported bipedal robot, depicted in Fig.~\ref{fig.robot}.
This physical robot allowed us to apply the same gait optimization algorithm in different joint configurations.
We performed experiments both in simulation using OpenAI gym and in the real world on the physical robot, and recorded the performance of the different ankle joint configurations.

Related research in bipedal robots have implemented both toe and heel joints for bipedal walkers \cite{steinars,planningwalking}.
They have shown that it increases the support area during double support and improved effectiveness in terms of energy consumption and total torque requirements.
Other studies in gaits and walking patterns remove these joints in their simulations and implementations to simplify the problem, as seen in \cite{hub}.

In this paper, we aim to investigate how adding an ankle joint to a supported bipedal robot improves performance when the gait of the robot is optimized.
We chose to use evolutionary algorithms for the gait optimization as they are widely used in legged robots and well suited for the relatively limited evaluation budget in hardware.
We have designed a robot leg with an actively controlled ankle joint and compared it to a leg with both a static ankle and no ankle joint at all.



Our experiments demonstrate the benefit of having an ankle join in a bipedal robot when it is subjected to gait optimization through evolutionary algorithms.
The robot reaches a much higher speed while walking with an active ankle joint than with a static ankle or no ankle.

\section {Background}

Bipedal humanoid robots are becoming more popular and a range of different robots are now available, from simple toys to advanced robots for research and manufacturing~\cite{saeedvand2019comprehensive}.
Two legged robots can traverse more challenging terrains than their wheeled counterparts, and are a versatile platform that suits many applications.
One of the biggest challenges with bipedal robots is achieving high robustness to different walking surfaces and terrain disturbances~\cite{xie2020review}.
Bipedal walkers also face unique challenges when it comes to sensing and navigating complex environments~\cite{ortega2019review}.

Today, research in the field of bipedal walkers focus more on new and improved control methods than new mechanisms and novel morphologies~\cite{xie2020review}, and a range of different machine learning techniques are used~\cite{wang2012machine}.
Many researchers draw inspiration from animals or humans~\cite{ames2014human,jiang2012outputs}, but directly comparing walking behavior and performance is very challenging, and no widely used standard benchmarks are  available~\cite{torricelli2020benchmarking}.
Running is considered especially hard to copy, though some have achieved good results by modelling the same spring-damper quality found in human legs~\cite{athlete}.
Model-based approaches have successfully been used to better understand and analyze the behavior and performance of biped robots~\cite{kumar2020review}.
Achieving high stability is a challenge, but machine learning approaches like supervised learning has been shown to be able to produce stable locomotion on underactuated bipedal robots~\cite{da2017supervised}.
Evolutionary algorithms have been shown to be particularly useful~\cite{wahde2002brief}, and examples demonstrate optimization of bipedal locomotion in both simulation~\cite{arakawa1996natural} and hardware~\cite{wolff2001evolution}.

Optimizing the control of legged robots can have a great effect on performance, but changing the body of a robot can also have a large effect on behavior and performance~\cite{nygaard18exploring}.
The design of bipedal robots can vary a lot, and some use serial, parallel or even hybrid mechanical mechanisms~\cite{gim2018design}.
There are examples of using evolutionary techniques and other machine learning approaches to optimize the morphology of biped robots, but these are mostly done in an offline fashion on a simulated robot~\cite{paul2001road}.
Working on a virtual robot in simulation bypasses many of the challenges associated with real-world robots~\cite{nygaard2019experiences}, though the difference between the behavior and performance in simulation and reality makes the results far less effective and applicable for a physical robot~\cite{jakobi1995noise}.
There are examples of previous work demonstrating adaptation of legged robot morphology to the robot's operating environment through evolutionary techniques, both in an off-line~\cite{nygaard21environmental} and on-line fashion~\cite{nygaard21real-world}.
These, however, typically make smaller changes to the morphology at a time, and few studies with large differences in morphology has been conducted on legged robots.

The design possibilities with bipedal robots is almost endless, and creative solutions going beyond traditional legged robot design are becoming more common.
Some add wheels~\cite{klemm2019ascento}, although that comes with a wide range of new challenges~\cite{chen2020underactuated}.
Other ways of increasing the maneuverability and stability of a bipedal walker is to add a tail~\cite{rone2018maneuvering}.
Some even add novel mechanical elements like thrusters to achieve unusual characteristics and capabilities~\cite{harpy}.


\section{Implementation}

\subsection{The robot}

\begin{figure}
    \centering
    \begin{subfigure}[b]{0.25\textwidth}
        \includegraphics[width=\textwidth]{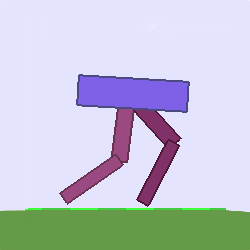}
        \caption{No Foot}
    \end{subfigure}
    \begin{subfigure}[b]{0.25\textwidth}
        \includegraphics[width=\textwidth]{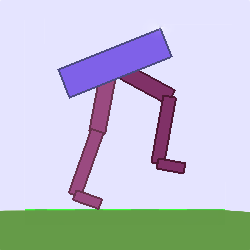}
        \caption{Static}
    \end{subfigure}
    \begin{subfigure}[b]{0.25\textwidth}
        \includegraphics[width=\textwidth]{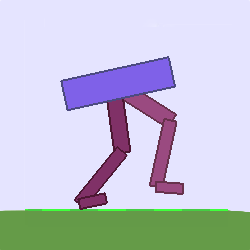}
        \caption{Active}
    \end{subfigure}

    \begin{subfigure}[b]{0.25\textwidth}
        \includegraphics[width=\textwidth]{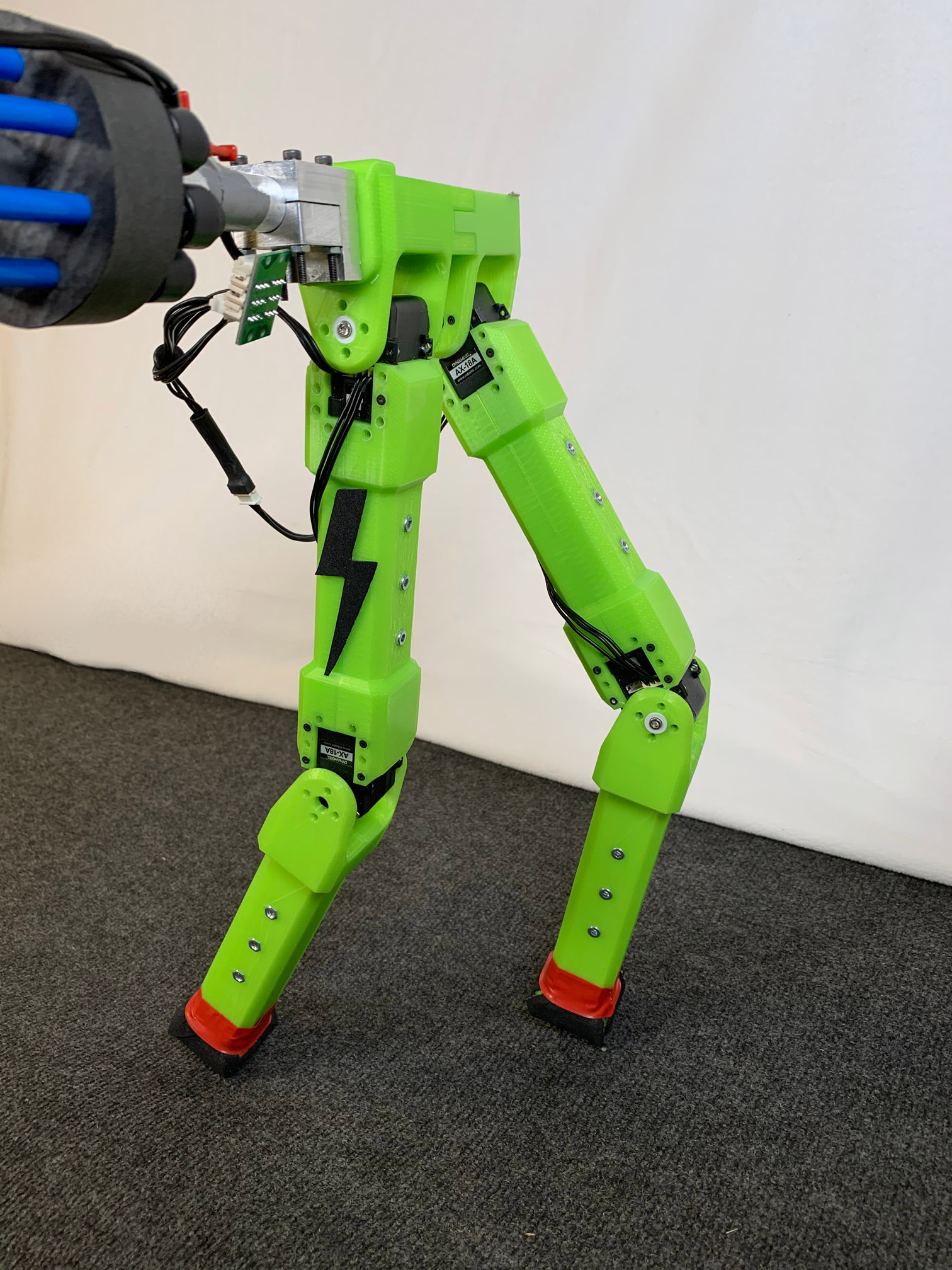}
        \caption{No Foot}
    \end{subfigure}
    \begin{subfigure}[b]{0.25\textwidth}
        \includegraphics[width=\textwidth]{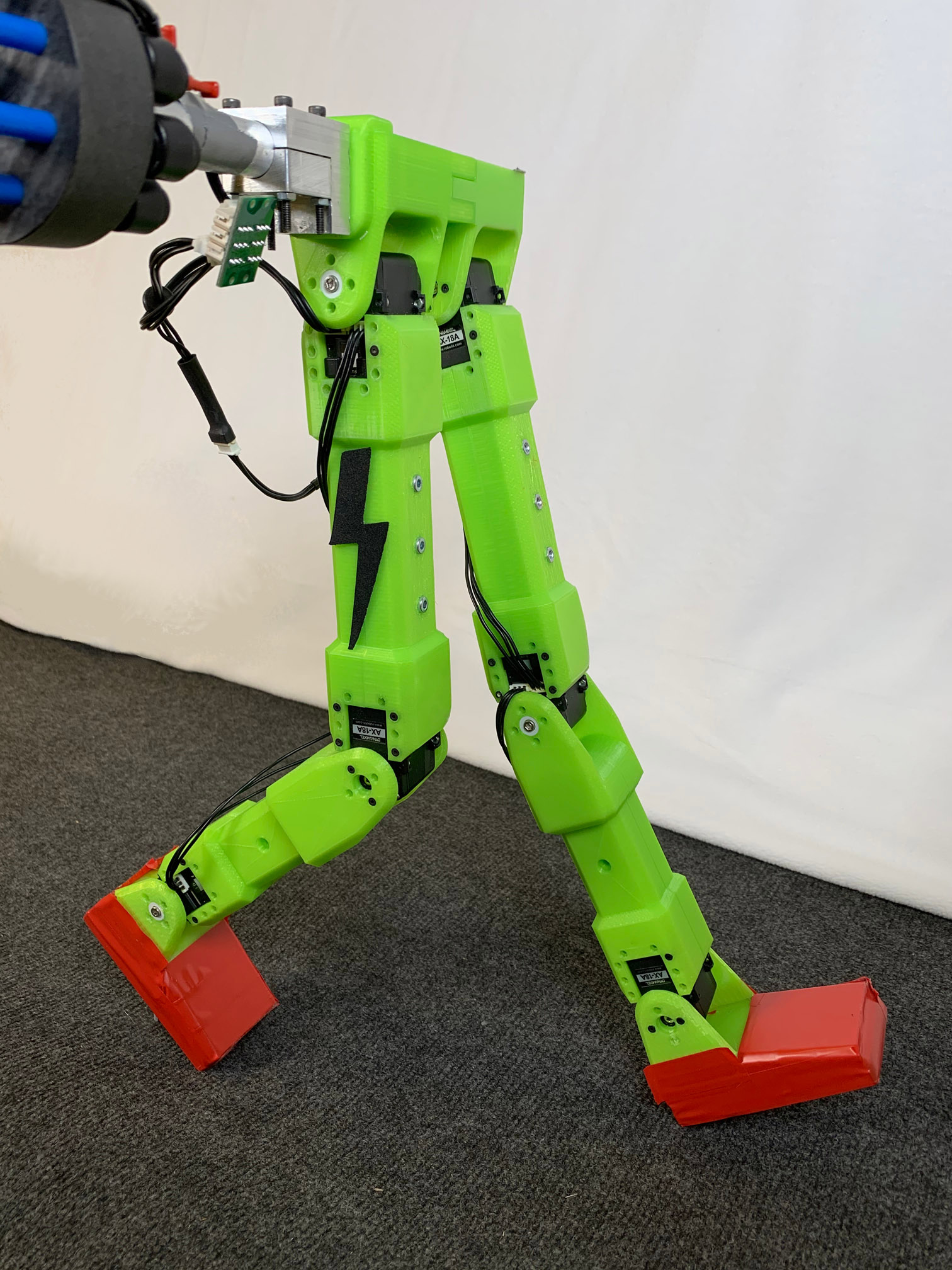}
        \caption{Static}
    \end{subfigure}
    \begin{subfigure}[b]{0.25\textwidth}
        \includegraphics[width=\textwidth]{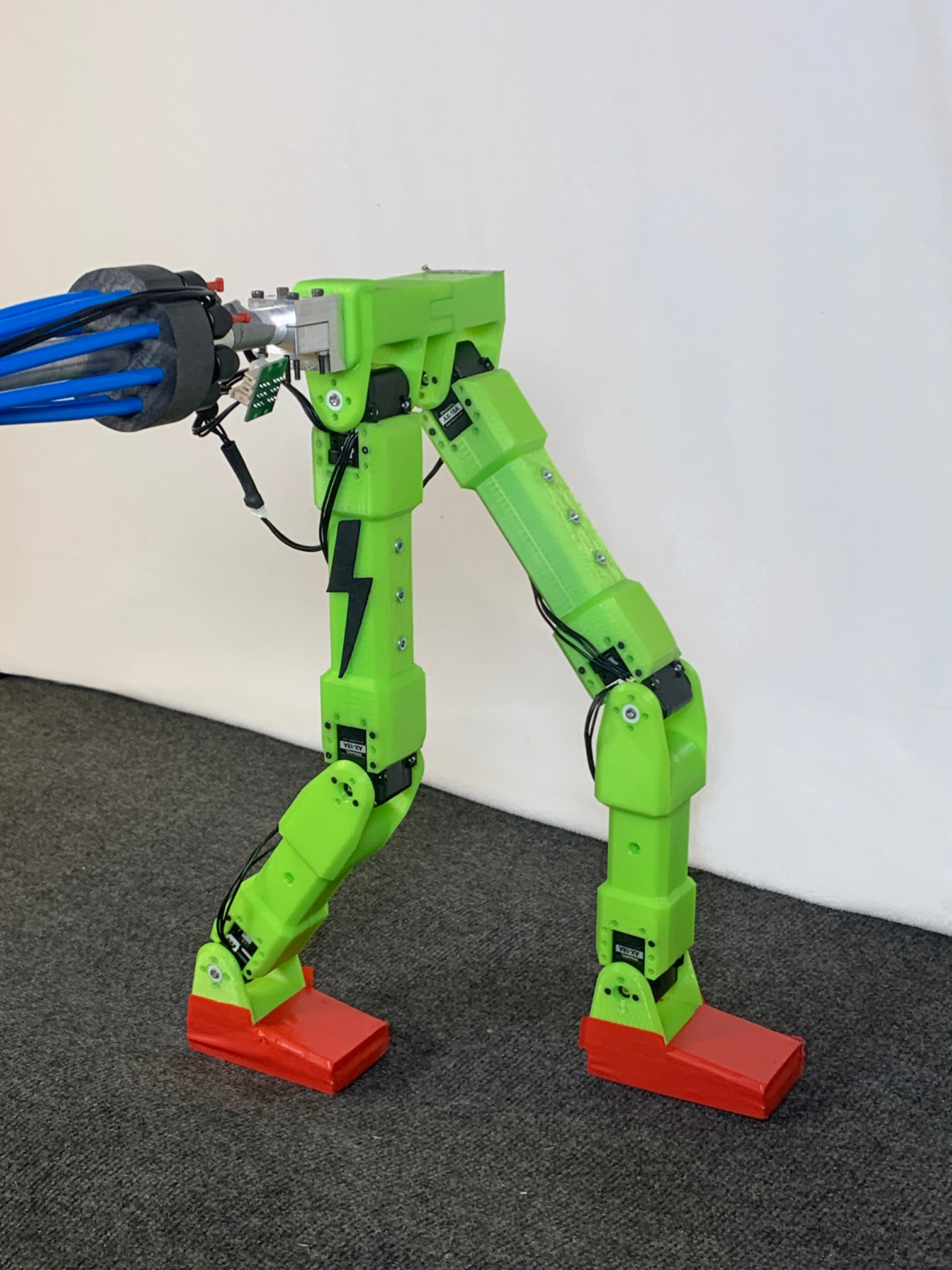}
        \caption{Active}
    \end{subfigure}
    \caption{Morphologies in hardware}
    \label{fig.hardware}
    \vspace{-5mm}
\end{figure}

\begin{figure}[t]
  \centering

  \begin{subfigure}[b]{.44\textwidth}
      \vspace{3mm}
      \hspace{-2mm}
      \includegraphics[width=6.0cm]{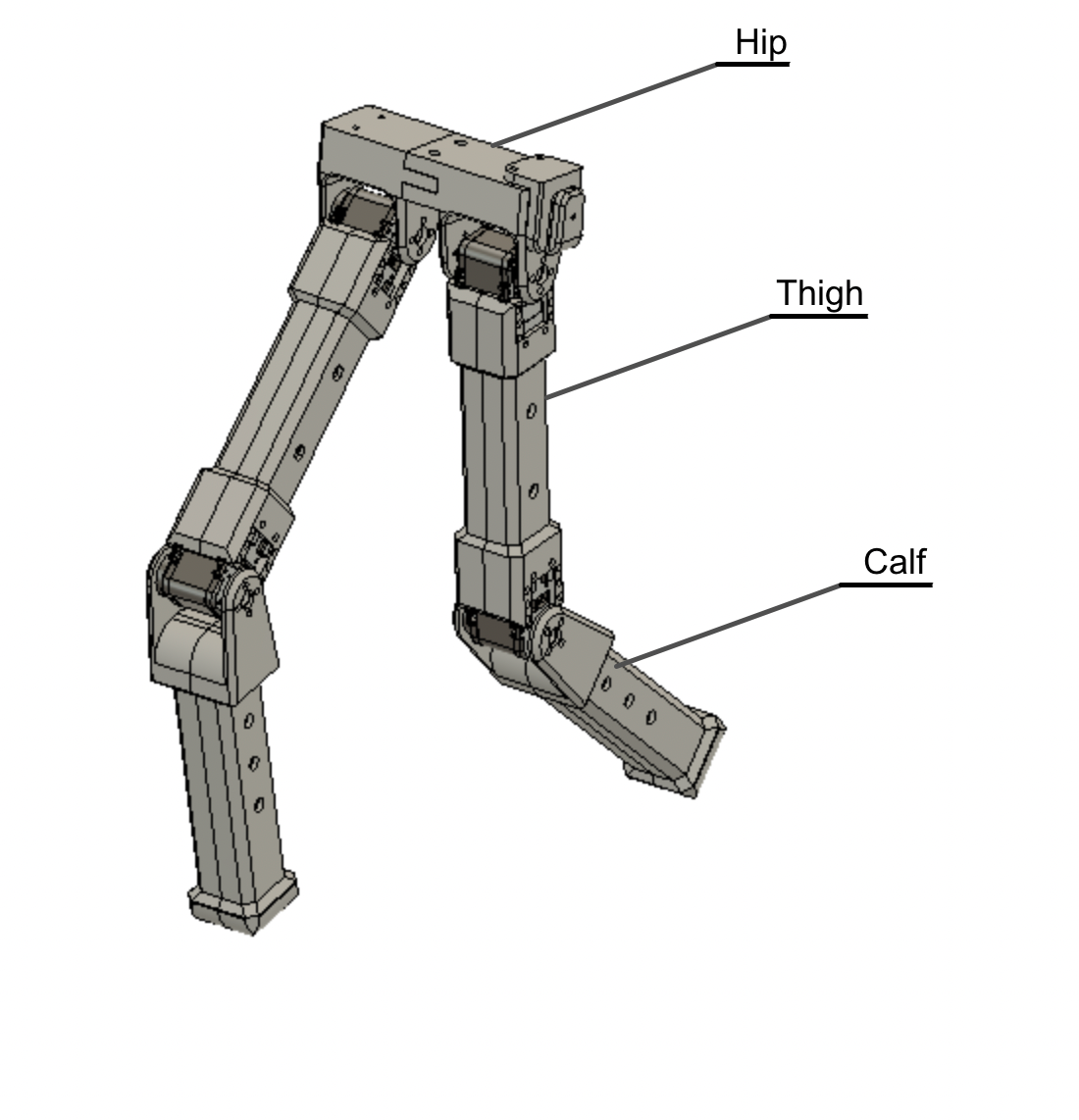}
      \caption{Design without an ankle -- referred to as the noFoot morphology.}
      \label{fig.names_noFoot}
  \end{subfigure}%
  \hspace{5mm}%
  \begin{subfigure}[b]{.48\textwidth}
      \includegraphics[width=6.0cm]{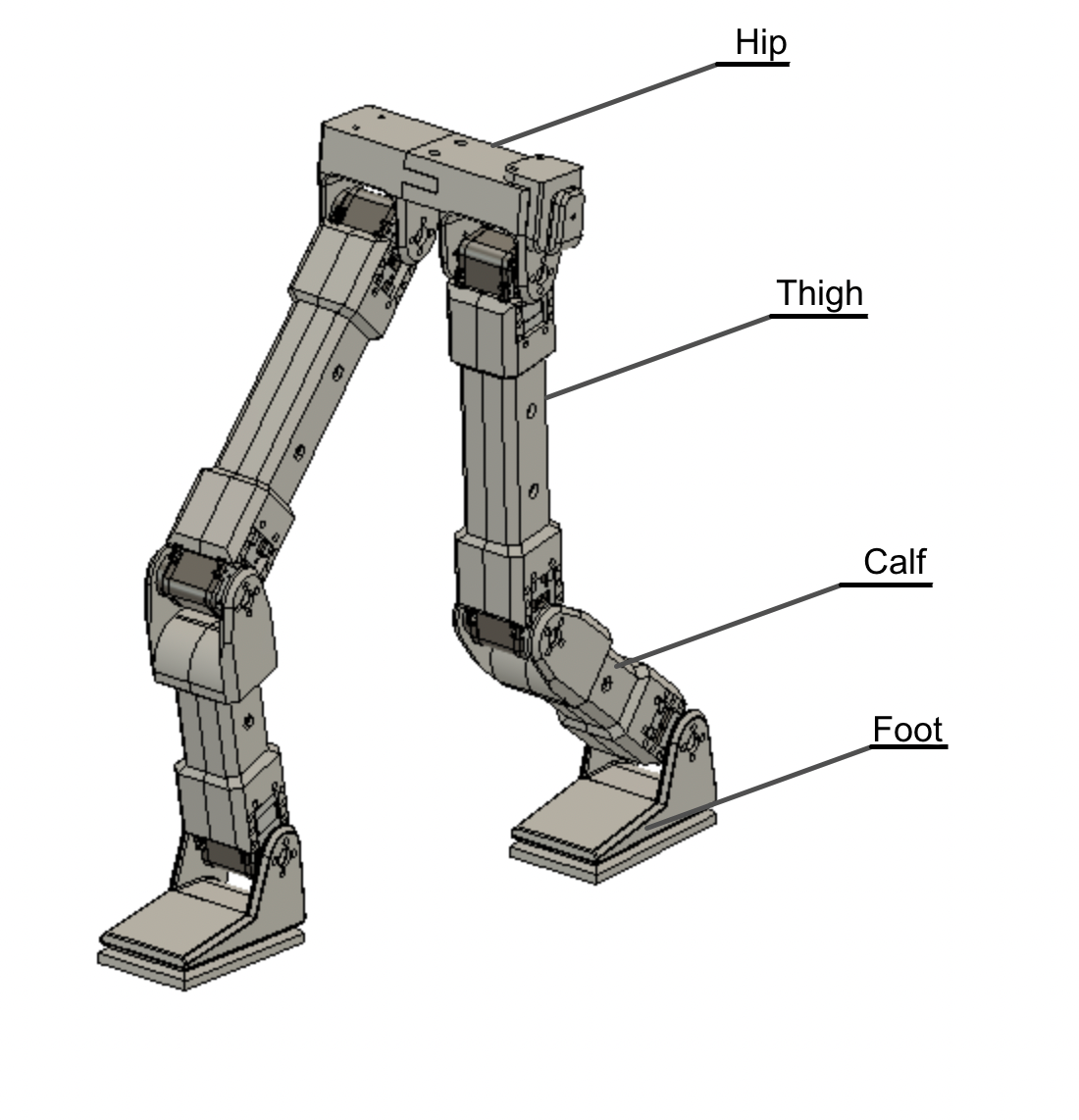}
      \caption{Design with an ankle used both with active control and at an static angle.}
      \label{fig.names_active}
  \end{subfigure}

  \caption{Diagrams of the physical robot used in the real-world experiments.}
  \label{fig.names}
  \vspace{-5mm}
\end{figure}

The physical robot consists of a central body connected to the two legs as seen in Fig.~\ref{fig.names}.
The simplest configuration only has two joints, and is referred to as the "no foot" morphology.
The hip joint connects the body and the thigh, and the knee joint connects the thigh and the calf.
The other configurations include an ankle joint between the calf and the foot.
The "static" morphology keeps this joint fixed at 90 degrees, and it does not move while walking.
This angle has previously been shown to be optimal in similar robots~\cite{claes}.
The "active" morphology uses the ankle joint as a functional part of the leg while walking.

The body is connected to a freely rotating hub supplying power and control.
This rigid connection ensures that the body of the robot is kept straight during walking, allowing us to disregard balancing.
Reducing the complexity of the problem like this is commonly done in legged robotics~\cite{harpy}, and simplifies both the design of the robot and the locomotion task the robot attempts to solve.
It also makes experiments simpler by increasing the repeatability while reducing the risk of damage to the robot or experimental setup.

\subsection{Gait controller}


One step consists of two different states, a lift state and a ground state.
In the ground state, the leg moves straight along the ground between two points (touchdown and liftoff in Fig.~\ref{fig.spline}).
In the lift state, the feet follow a curve determined by the liftoff\_control and touchdown\_control points.
All points are represented by their cartesian coordinates relative to the center of the hip joint (origo in Fig ~\ref{fig.spline}).
The spline forms a bezier curve.
The two legs are offset 180 degrees out of phase to ensure one leg always touches the ground.
The gait controller works in cartesian space, and inverse kinematics are used to convert the positions to angles for the individual joints.

Controlling the morphology with the active ankle joint adds three extra parameters to the gait controller.
The foot was kept parallel to the ground in the ground state, but made a kick in the beginning of the lift phase and returned to parallel at the end.
The kick was described by three parameters using the following function


\begin{equation}
    \text{amount} *(\frac{\text{speed}-\text{offset}}{3})^2 *  e^{1-(\frac{\text{speed}-\text{offset}}{3})^2}
\end{equation}

This function was chosen as a naive and novel heuristic to approximate human use of an ankle through initial experimentation.
It reaches its maximum quickly and returns to the ground state over a predictable time period.
The amount ($ankle_extension_amount$) determines the maximum extension of the joint, speed ($ankle_extension_speed$) determines how fast the maximum extension is reached, and offset ($ankle_extension_offset$) decides when in the the gait cycle the activation pattern starts.

\begin{table}[t]
    \centering
    \caption{Gait parameters and ranges.}
    \label{table.params}
    \begin{tabular}{|l|l|l|}
      \hline
      \bfseries Category               & \bfseries Name             & \bfseries Values \\ \hline
        \multirow{2}{60pt}{Spline shape} & \textit{touch\_down\_x}      & [ 0, 235mm]   \\ \cline{2-3}
                                       & \textit{touch\_down\_control\_x}   & [   0,  235mm]    \\ \cline{2-3}
                                       & \textit{touch\_down\_control\_y}   & [   285mm,  325mm]    \\ \cline{2-3}
                                       & \textit{lift\_off\_x}   & [-235mm, 0mm]    \\ \cline{2-3}
                                       & \textit{lift\_off\_control\_x}   & [   -325mm,  0mm]    \\ \cline{2-3}
                                       & \textit{lift\_off\_control\_y}   & [   285,  325mm]    \\ \cline{2-3}
      \hline
      \multirow{2}{60pt}{Ankle activation}  & \textit{ankle\_extension\_amount}   & [0, 90\degree]   \\ \cline{2-3}
                                         & \textit{ankle\_extension\_speed}   & [ 0, 270\degree /s]    \\ \cline{2-3}
                                       & \textit{ankle\_extension\_offset}   & [0\%, 100\%]    \\ \cline{2-3}
    
      \hline
    \end{tabular}
\end{table}

\subsection{Simulation}

To simulate our solutions, we used the BipedalWalker environment from Open AI gym~\cite{brockman2016openai}.
Although this is originally intended for reinforcement learning purposes, the ease of set-up and wide-spread use in the field makes it a good option.
The framework originally takes joint velocities as inputs to the motors, but a new joint control module was added that instead takes in joint angles.
This was required since we wish to use our solutions on a physical robot that has joint position control.
The density of the body was set to 5 to approximate the physical robot.

A reward function taking into account forward movement, the movement of the joints, as well as the angle of the robot body was used due to the framework being designed for reward functions from reinforcement learning. The full details are given in equation~\ref{eq.reward}.

\begin{equation}
    \Delta pos - 3.5*10^{-4}*\Delta joints -  5 *angle_{body} - \frac{1}{2}
    \label{eq.reward}
\end{equation}

The fitness of an individual is defined as the sum of these rewards over 12 periods.
This was done to reduce the noise from real-world measurements and increase the repeatability of evaluations.



\subsection{Real-world evaluation}

The body and legs were designed and 3D printed in PLA material.
Dynamixel AX-18 smart servos were used as the joint actuators.
These are position controlled, and can therefore be controlled similarly to the simulation model.



To simplify the mechanical complexity of the robot, the legs were mounted at the hip to a beam that rotated around an axis as depicted in Fig. \ref{fig.robot}.
This hub setup ensured the robot walked in a circle.
It also meant we would not have to worry about sideways balance, making it close to the 2D simulation environment.


\subsection{Evolutionary setup}

A simple evolutionary algorithm was used to optimize the gait of the robot. Table~\ref{table.evoParams} shows the evolutionary parameters used in the experiment.

\begin{figure}[htb]
    \centering
    \includegraphics[width=11cm]{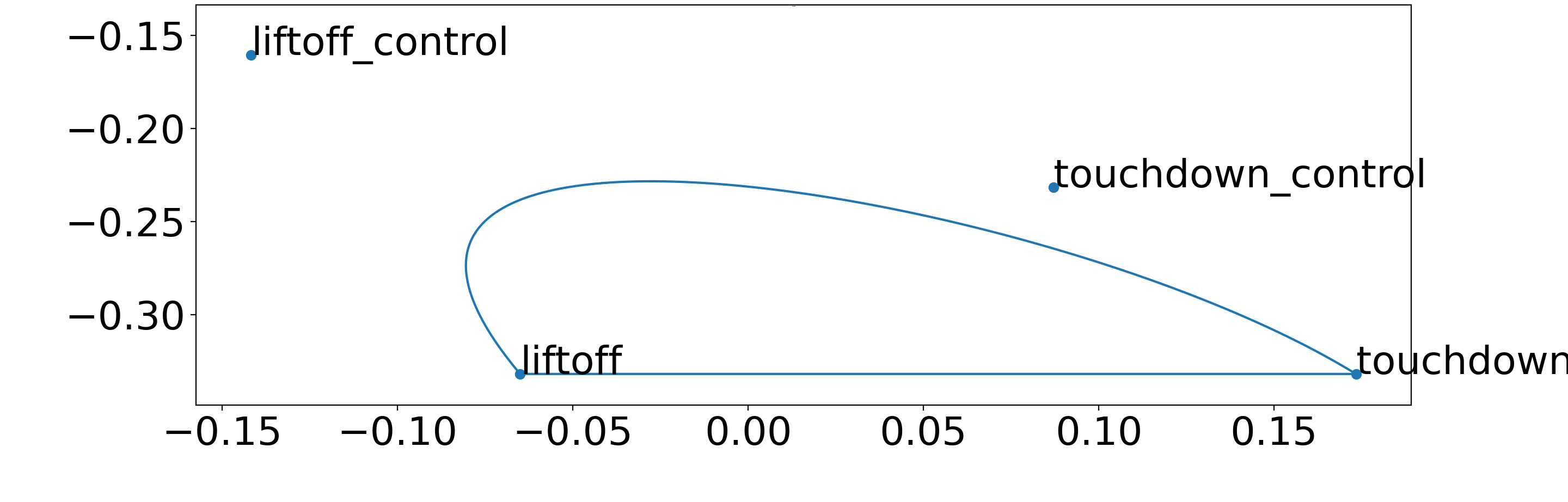}
    \caption{Visualization of the parameters in the gene deciding the control points for the spline}
    \label{fig.spline}
\end{figure}

The genotypes were a set of numbers between 0 and 1, that describes the walking gait, as seen in Fig.~\ref{fig.spline}. 
When training on the active ankle joint morphology, the genotype consisted of an undefined number of extra parameters.
They were used to derive a function describing the movement of the ankle, and decided by an input parameter. 

Initialization was done by randomizing all genotype values.
25 percent of each new generation was chosen at random and 25 percent was chosen based on novelty (the euclidean distance of the parameters compared to the rest).
The absolute best solution, together with a fitness-proportionate selection of the remaining solutions were used as the next generation.
This ensures high exploration in the search.

Uniform mutation was used along with PMX recombination.
The genotype was clamped to ensure only valid values were kept.

\begin{table}[htb]
  \centering
  \caption{Parameters for the evolutionary experiments}
  \label{table.evoParams}
  \begin{tabular}{|l|l|}
    \hline
    \bfseries Name & \bfseries Value \\
    \hline
    Recombination & PMX \\
    \hline
    Parent selection & Fitness proportionate selection \\
    \hline
    Novelty selection & Euclidean distance of parameters \\
    \hline
    Algorithm runs & Once per mutation probability and generation combo\\ \cline{2-2}
    \hline
    \multirow{2}{60pt}{Mutation}    
                                    & Type: Uniform distribution over [$-prob^2$, $prob^2$) \\  \cline{2-2}
                                    & Probability: 0.3, 0.5, 0.7 \\ \cline{2-2}
    \hline
    \multirow{2}{60pt}{Evaluations} & Population: 256 \\  \cline{2-2}
                                    & Generations: 50 \\ \cline{2-2}
                        \cline{2-2}
                                    & Evolutionary runs per joint configuration: 9 \\
    \hline
    \multirow{2}{60pt}{Next generation} & 1/4 Chosen parents - mutated \\ \cline{2-2}
                                        & 1/4 Generated children - mutated \\ \cline{2-2}
                                        & 1/4 Random initialized solutions \\ \cline{2-2}
                                        & 1/4 Solutions based on novelty - mutated \\
    \hline
  \end{tabular}
\end{table}

\section{Experiments and Results}
The role of the ankle joint is first investigated through experiments in simulation before they are also tested on a physical robot in the real world.

\subsection{Optimization in simulation}

To optimize the gait in simulation, the evolutionary algorithm was run with a population size of 256 for 50 generations.
Initial experimentation showed this to give a good balance between exploration and exploitation.
Separate evolutionary runs were done for all three joint configurations.

\begin{figure}[htb]
  \centering
  \includegraphics[width=0.8\linewidth]{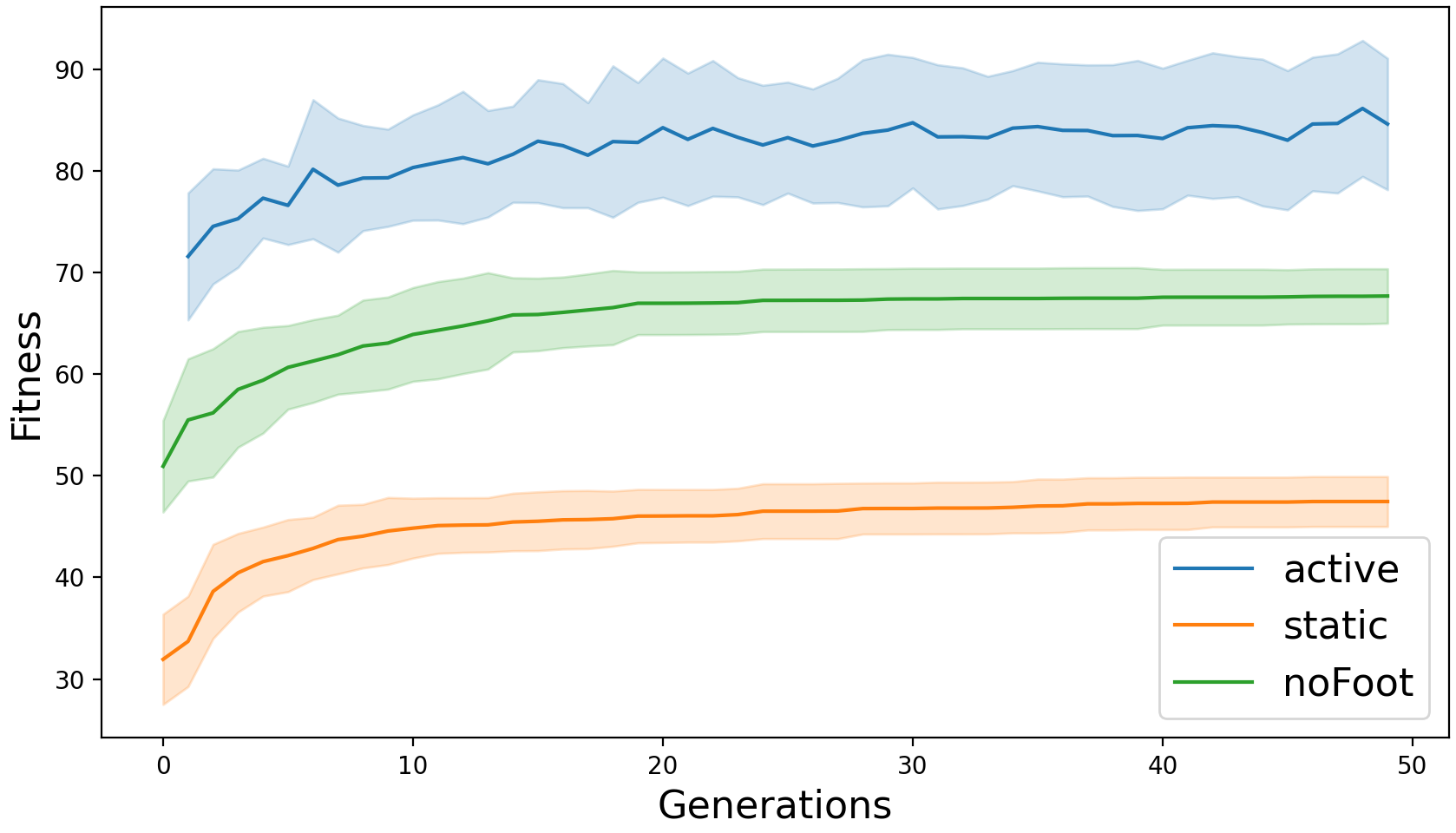}
  \caption{Average reward $\pm$ std.dev over the generations without balance modelled}
    \label{fig.results}
\end{figure}

Fig.~\ref{fig.results} shows the fitness of the evolved individuals throughout the evolutionary runs.
We see that the static ankle configuration does significantly worse than the others, ending with an average reward of about 47.
The nofoot configuration ends up in the middle with an average fitness of about 68, while the active joint configuration does best, ending at an average fitness of about 85.
The evolution of both the static and nofoot configurations converge in under half of the generations, while the fitness of the active joint configuration is less stable overall.

Table~\ref{table.results} shows the average values for each of the three joint configurations.

\begin{table}[htb]
    \centering
    \caption{Mean and standard deviation of the individuals in generation 50 for all morphologies}
    \label{table.results}
    \begin{tabular}{|l|ll|ll|ll|}
      \hline
           & \multicolumn{2}{l|}{Active}     & \multicolumn{2}{l|}{static}      & \multicolumn{2}{l|}{Nofoot}                          \\ \cline{2-7}
           & mean & std & mean & std  & \multicolumn{1}{l|}{mean} & \multicolumn{1}{l|}{std} \\ \hline
    Fitness & 84.61 & 6.47 & 47.47                     & 2.46 & 67.69                     & 2.68                     \\
    touch\_down\_x      & 0.88  & 0.22  & 0.35                      & 0.23 & 0.80                      & 0.23                     \\
    lift\_off\_x      & 0.75  & 0.28 & 0.83                      & 0.2  & 0.77                      & 0.19                     \\
    touchdown\_control\_x      & 0.2   & 0.13 & 0.21                      & 0.2  & 0.23                      & 0.28                     \\
    lift\_off\_control\_x      & 0.82  & 0.18 & 0.84                      & 0.19 & 0.75                      & 0.26                     \\
    lift\_off\_control\_y      & 0.3   & 0.37 & 0.30                      & 0.29 & 0.16                      & 0.22                     \\
    touchdown\_control\_y      & 0.18  & 0.2 & 0.16                      & 0.26 & 0.09                      & 0.2                      \\
    ankle\_extension\_amount      & 0.52  & 0.42  & -                         & -    & -                         & -                        \\
    ankle\_extension\_speed      & 0.54  & 0.35 & -                         & -    & -                         & -                        \\
    ankle\_extension\_offset     & 0.65  & 0.41 & -                         & -    & -                         & -
    \\
    \hline
    \end{tabular}

\end{table}

\subsection{Real-world experiments}

The best solution from evolution on each joint configuration was selected for testing on the physical robot.
They were allowed to walk for a short time on the hub and the distance covered and time taken was recorded to calculate the speed of the robot.

Table~\ref{table.selected} shows the results after running the best solutions on the physical robot.
We see that the noFoot configuration has the worst velocity at 0.03m/s.
The active configuration has a speed of 0.08m/s, while the static configuration outperforms it and achieves a speed of 0.24m/s.
This is in contrast to the simulated experiment, where the active configuration was the fastest one.

\begin{table}
    \centering
    \caption{Measured performance of the different joint configurations in the real world.}
    \label{table.selected}
    \begin{tabular}{|l|l|l|l|}
      \hline
               & Distance (m)& Time(s) & Avg. velocity(m/s) \\ \hline
        Active &       3.46  &   40.86 &               0.08 \\
        Static &       6.91  &   28.04 &               0.24 \\
        noFoot &       3.46  &  117.63 &               0.03 \\
        \hline
    \end{tabular}
\end{table}

\FloatBarrier

\section{Discussion}

The fact that the active joint configuration in simulation performs so much better than both static and nofoot configurations supports the notion that the ankle joint could be key to better performance for bipedal robots.

In hardware, however, we see that the static configuration performs the best, with the active configuration second.
This difference might simply be due to the reality gap -- the inaccuracies in the simulator causing it to give wrong behavior and performance measures.
This means that the controller optimized for the static configuration in simulation transferred well to the real robot, while the one optimized for the active configuration did not.
The fact that there is such a large difference between nofoot and static in the real world is most likely due to the geometry of the end of the foot.
Differences in the evolved control between the nofoot and static joint configuration could also contribute, but exploring this further is beyond the scope of this paper.

The evolutionary optimization of the nofoot and static configuration quickly converged to stable values, while the active configuration took much more time and was more unstable throughout.
The active configuration does have three more parameters to search through, which explains the additional time to convergence.
Why it is so much more unstable is harder to say, and further experiments and an analysis of the genotype of the resulting individuals might yield more information.

\section{Conclusion}

In this paper we investigated how adding an ankle joint to a supported bipedal robot improved performance when the gait of the robot was optimized.
Experiments in simulation showed that evolving a legged robot with an active ankle joint resulted in significantly better performance than without it.
We also performed preliminary testing on physical prototypes of three different ankle joint configurations.
There we saw that a static ankle joint performed the best, outperforming the active joint as well as a leg configuration without a full foot.
Our work in simulation directly showed the benefit of having an active leg joint when gait optimization is done on bipedal robots, and our preliminary testing in the real world indicated that this could transfer well to the real world.

\section{Future work}

We only did gait optimization in simulation, and this made it more challenging to analyze the results from the hardware experiments.
Instead doing the optimization directly in hardware would give a better estimate of the benefit of the ankle joints under real-world conditions.
We used a high level gait controller in cartesian space.
A more low level controller would allow a more diverse set of walking patterns that might be better suited to the different leg configurations than our more limited controllers were.
Our analysis was done the whole generations produced by evolution.
Instead looking at the behavior at the individual level could yield additional insights into the effects from different ankle joint configurations.
We optimized performance by looking at the speed of the individuals.
There are many alternative performance measures that could give interesting results.
These include energy usage or stability.

\bibliographystyle{splncs04}
\bibliography{bibliography}

\end{document}